\documentclass{llncs}
\usepackage[utf8]{inputenc}
\usepackage{amsmath}
\usepackage{algorithmic}
\usepackage{algorithm}
\usepackage{graphicx}
\usepackage{subfig}

\title{Continuous On-line Evolution of Agent Behaviours with Cartesian Genetic Programming}

\author{Davide Nunes, Luis Antunes}
\institute{LabMAg / Universidade de Lisboa \\ \email{ \{dnunes,xarax\}@fc.ul.pt }}

\begin{document}

\maketitle

\begin{abstract}
Evolutionary Computation has been successfully used to synthesise controllers for embodied agents and multi-agent systems in general. Notwithstanding this, continuous on-line adaptation by the means of evolutionary algorithms is still under-explored, especially outside the evolutionary robotics domain. In this paper, we present an on-line evolutionary programming algorithm that searches in the agent design space for the appropriate behavioural policies to cope with the underlying environment. We discuss the current problems of continuous agent adaptation, present our on-line evolution testbed for evolutionary simulation.
\end{abstract}

\section{Introduction}
In on-line scenarios, adaptive agents strive to find behaviour policies to cope with their environment while interacting with and learn about their surroundings. Adaptive agents can be designed using different paradigms, notably reinforcement learning, plan-based or supervised learning approaches. Supervised learning techniques require prior knowledge about the environment and use labelled training examples to construct appropriate behaviour policies. Using supervised learning techniques to synthesise agent behaviour policies is more appropriate for off-line learning scenarios where a proper assessment can be made about the state of the environment and the agents' behaviour can be properly evaluated in the light of the simulation results. There are two major problems with this approach: the first being that synthesising a successful adaptive controller is as dependent on scenario modeller as the plan-based approaches are on the agent designer. Furthermore, real-world environments can be dynamic and prior knowledge about all the aspects of the environment might not be available. We then claim that the development of on-line learning approaches is fundamental to construct truly autonomous adaptive behaviours.

Reinforcement learning, can be applied to problems for which significant domain knowledge does not exist. Its goal is to allow for the learning of optimal behaviour policies that produce the greatest cumulative reward over time. Not only has reinforcement learning made progress through direct interaction with complex sequential task environments, but also can shed some light on how to adapt traditionally off-line learning techniques to learn in a continuous manner in unkown and possibly dynamic environments. 

Other promising approaches come from the field of evolutionary computation. Evolutionary approaches such as genetic programming or neuro-evolution have shown promises in large and partially observable environments \cite{Gomez2006} but have been majorly targeted at off-line learning problems using supervised learning and simulated environments. One example is the synthesis of controllers for robots (embodied agents) that test behaviour policies in a simulated environment. The controllers are adapted using different what-if scenarios and then attached to real robots that are expected to perform adequately in the real-world. Some advances have been made in the usage of evolutionary algorithms to learn in an on-line fashion, this is, the robot is expected to adapt while learning and interacting with the unknown environment \cite{nordin1997line}. 

We are interested in the on-line evolutionary approach as a framework for intelligent agent design outside the domain of evolutionary robotics. We rely on evolutionary programming techniques to search for controllers that follow certain motivations. These motivations can be designed to drive the synthesis of utility-centric behaviour or allow for more complex social dynamics by pushing agent behaviour towards imitation, competition, cooperation, or simply \textit{ad-hoc motivations}. 

To build our genetic programming controllers, we use representation developed for Cartesian Genetic Programming (CGP) \cite{miller2000}. In this representation, a graph is encoded as a string of integer or real-valued numbers \cite{Clegg2007}. Among the multiple interesting features of this representation such as the support for multiple output and sub-graph activation or deactivation, it also allows for different search methods to be easily applied to the program synthesis process.

In this paper we explore an evolutionary algorithm for on-line learning based on evolutionary strategies and CGP. We study the influence of two different behaviour selection policies. These selection mechanisms dictate which programs are to be tested in the environment on each evaluation cycle and have a considerable impact in the agent capacity perform better or maintain behaviour diversity. The article is organised as follows. In the following section, we will present and discuss the problems found in on-line learning scenarios. Section 3 describes our simulation experimental design, target goals and used metrics. Section 4 presents and discusses the simulation results. Finally, section 5 presents our conclusions and considerations for future work. The relevant literature will be presented as we progress.

\section{On-line Evolution of Behaviour}
Among the multiple problems of evolving controllers for multiple agents in an on-line fashion is the fact that the behaviour policy trial and error process might induce noise, especially if the search process is focused on other agents that evolve at different rhythms and are subjected to different environment contexts. Notwithstanding this, some literature suggests that sparse training data sets or probabilistic sampling in evolutionary algorithms can actually increase the speed towards the goals and escape local optima by maintaining highly diversified solutions in the population \cite{nordin1997line}. Also, evolutionary algorithms actually cope well with noisy evaluations, claim illustrated in \cite{Fitzpatrick1988}.

Other major problem we have face in on-line evolution of behaviour is the fact that contextual information is decisive for the success of adaptation. If an agent adopts a policy that proves highly rewarding in some situation, that success might be achieved for contextual reasons that may not replicate in a different situation or context. Moreover, we do intend that our agents are deployed in continuously changing environments, so it is a matter of time that the context changes and the successful policy must be adapted or even radically changed to face the new situation. 

To deal with the different evaluation conditions, the selection of which controller program to use in a given situation is a keystone of our approach. We test three different mechanisms. In the first, each program is exposed to the same number of tests but under different testing conditions (altered by the action of the agent over its environment). The second method is based on confidence intervals, estimated from the program evaluations over time, which allows for programs with better performance or more uncertainty to be selected and tested. Finally we combine these approaches with energy-based stabilisation, where agents maintain their current controller program running if a good performance is being achieve, and switch to other program (based on uniform probability or performance estimation). Section \ref{sec:online-evo-alg} describes the program selection mechanisms that were tested.

\section{Simulation Methodology, Model, and Metrics}
In this section we describe implementation of our simulation model for on-line behaviour evolution, the metrics used to produce our results, and describe the environment in which agents execute their behaviours.
 
\subsection{2D Fitness Landscape}
\label{sec:landscape}
\begin{figure}
     \centering
     \subfloat[]{
   \includegraphics[width=0.40\linewidth]{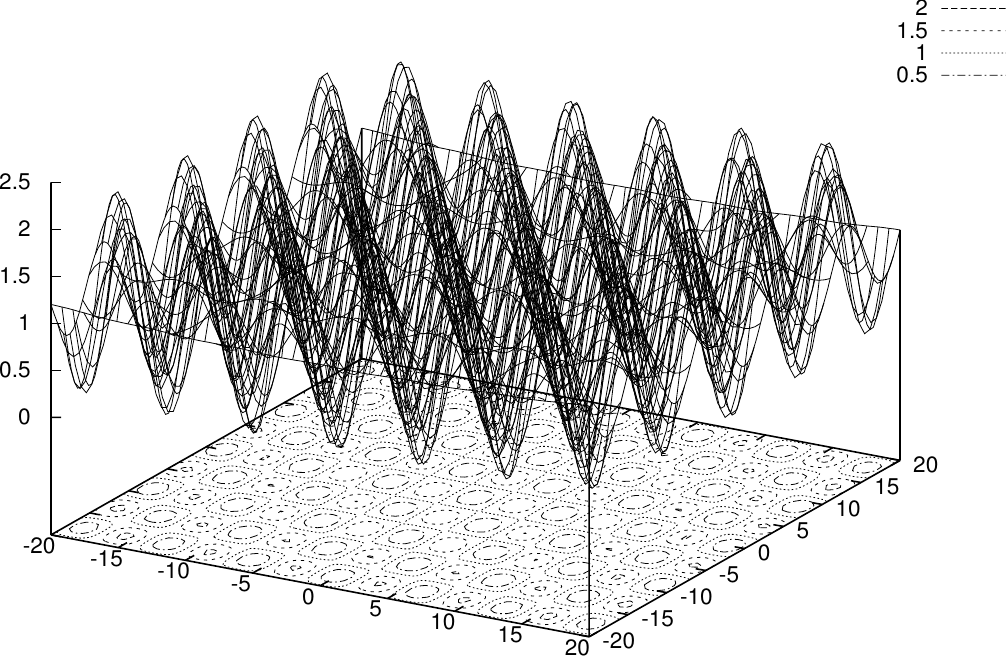}
     \label{fig:griwank}
     }
     \subfloat[]{
\includegraphics[width=0.40\linewidth]{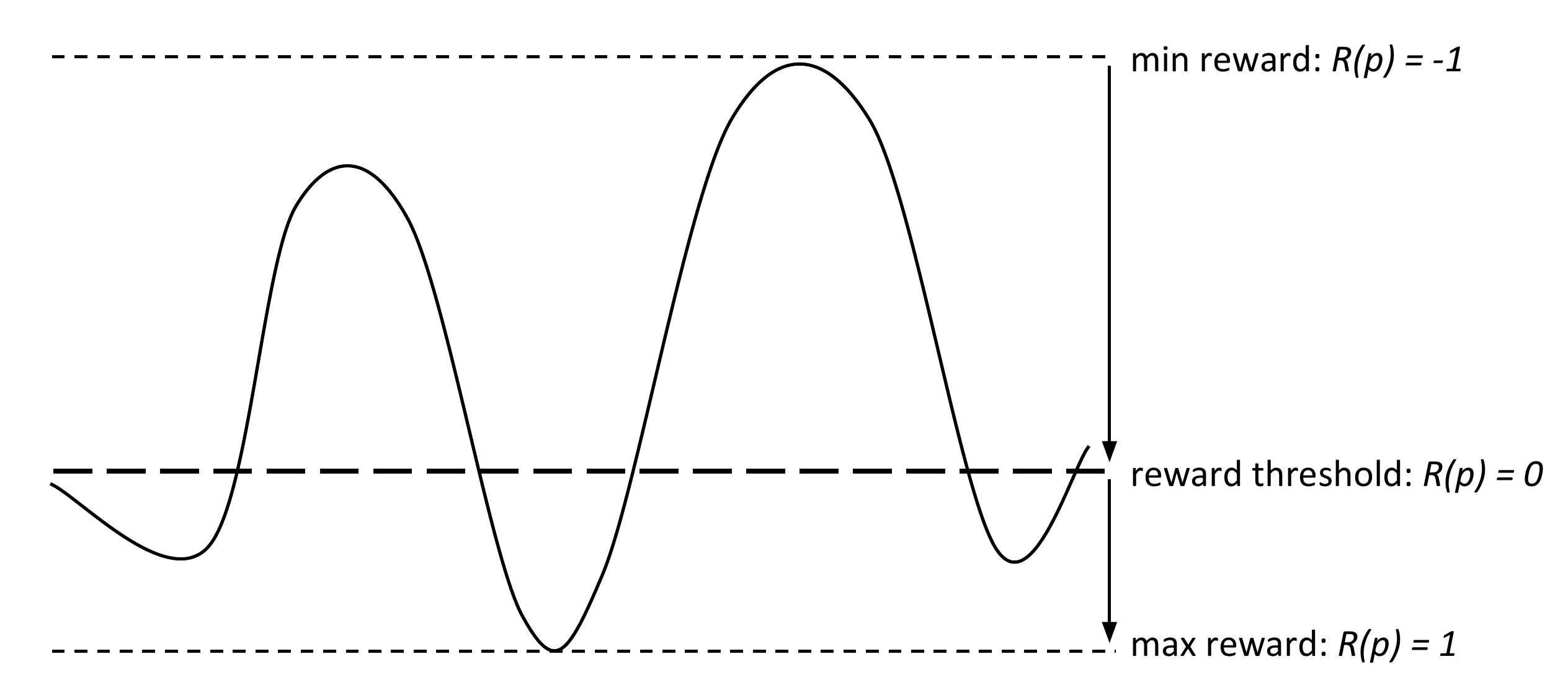}
   \label{fig:reward}
     }\\
     \begin{minipage}{0.8\linewidth}
     \caption{The graph for a two-dimensional Griewank function for the coordinates $x = y \in [-20,20]$ in (a) and the method used to convert the fitness landscape values into rewards with values $R(p) \in [-1,1]$.}
     \end{minipage}
\end{figure}

Our aim is to study the usage of evolutionary algorithms to produce adaptive agents that learn to behave in an on-line manner. We want to create increasingly more difficult scenarios where agents have to learn as they interact with the environment. In this first paper, we place the agents in a 2D toroidal world which is previously unknown to the agents. We assign a fitness value to the regions in this 2D world according to a pre-defined fitness function. In this case, we used a \textit{Griewank} function \cite{griewank1981} (see equation \ref{eq:griewank} and figure \ref{fig:griwank}). The Griewank function of order $n$ is defined by: 

\begin{equation}
\label{eq:griewank}
f_n(x_1,\dotsc,x_n) = 1 + \frac{1}{4000} \sum_{i=1}^{n} x^{2}_{i} - \prod_{i=1}^{n}\cos \left(\frac{x_i}{\sqrt{i}}\right)
\end{equation}    

The agents aim for learning behaviours that allow them to remain in highly valuable regions. Figure \ref{fig:reward} shows the mechanism used to compute reward values from the Griewank fitness landscape. The reward threshold is a convenient way to adjust how ward is to find energy in this environment.

In future work, we intend to progressively increase the complexity of this scenario by adding multiple concomitant fitness landscapes or using dynamic worlds where the reward values change over time. Other source for dynamics could also come from agents that modify their environment.

\subsection{Cartesian Genetic Programming}
Cartesian Genetic Programming (CGP) \cite{miller2000} is a type of genetic programming where programs are represented as feed-forward acyclic directed graphs. Typically, in this cartesian representation, the genotype is represented in the form of a list of integers. In this paper we used a real-based representation where each gene $g_i$ has a real value such that $g_i \in [0,1]$. This representation was proposed in \cite{Clegg2007} and has been developed to allow for the usage of the crossover operator, which typically hinders the performance of CGP when used with the integer value representation. 

Among the benefits of CGP is the fact that it allows for the implicit re-use of nodes in the directed graph. The number of nodes in the program (phenotype) is bounded and not all of the nodes encoded in the genotype have to be connected. This allows areas of the genotype to be inactive and have no influence on the phenotype. Miller identified this feature as being key in reducing bloat even when enormous genotypes are allowed \cite{Miller2006}. In tree-based GP models, most equally good phenotypes differ from one another in useless (bloated) code sections, and they will be strongly selected for when the average population fitness is high. In CGP however, the increased proportion of genetically different but phenotypically identical code is able to exist without harm.

In this paper, agents evolve programs that control how they move in the 2D environment. It maps the agent state in this case its coordinates and its current velocity in the $x$ and $y$ coordinates and outputs the value for the next velocity to be used.

\begin{figure}
\centering
\includegraphics[width=0.5\linewidth]{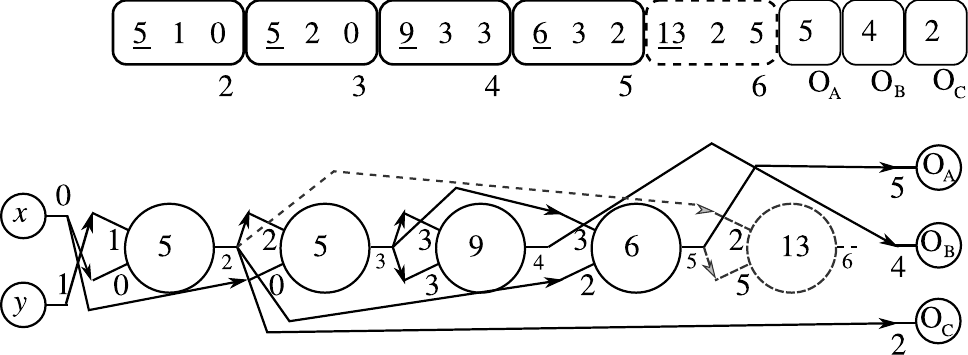}
\caption{CGP integer genome and its corresponding execution graph \cite{miller2000}.}
\label{fig:cgp}
\end{figure}

To construct the agent programs we use a function set composed of the following primitives: $F_{SET}$ = \{\textit{$+$}, \textit{$-$}, \textit{$\times$}, \textit{$/$}, \textit{sin}, \textit{comp} \}. We also use a constant set that can be used in the resulting program $C_{SET} = \{-0.5,0.5\}$. The $cmp$ primitive corresponds to the compare function which returns $-1$, $0$ or $1$ according to the comparison of its arguments.

\subsection{Evolutionary Strategies}
\label{sec:es}
The evolutionary algorithm used to search in the cartesian program design space is a $(1 + \lambda)$ evolutionary strategy \cite{schwefel1965} with $\lambda = 4$. In this algorithm, 1 parent has $\lambda$ offspring. The \textit{population size} is set to 5 as this configuration has proven efficient to evolve cartesian programs in most problems \cite{Miller2006}.

This evolutionary strategy is defined as follows. Let $f(g)$ be the fitness of a genotype $g$. Let $g_p$ be the parent genotype. The parent is mutated $\lambda$ times to generate $\lambda$ offspring $g_i$. Let $g_{p}^{'}$ be the new parent, selected from the population $\{g_p, g_i\}$ as follows:

$$
g_{p}^{'} =
\begin{cases}
g_i, & f(g_i) > f(g_j) \forall i \neq j \\
g_i, & f(g_i) = f(g_p) \text{, select lowest } i \\
g_p, & f(g_p) > f(g_i) \forall i \\
\end{cases}
$$

\noindent In CGP the genotype has a large number of inactive genes and mutations cause many genotypes to decode to the same phenotype resulting in identical fitness. The constant genetic change in genotypes of the same fitness has been shown to be a main reason for this strategy to be efficient in numerous problems.

\subsection{On-line evolutionary algorithm}
\label{sec:online-evo-alg}
Our main objective in this paper is to show how one can use genetic programming in general, and CGP in particular, to adapt and evolve a population of agents while they interact with the environment. Unlike off-line settings, we do not re-start the simulation in a per-generation basis restoring the conditions in which the agents are evaluated. In this case, each agent tests different behaviour policies (which map its current state to a set of actions to take) under different situations. 

In our approach, each agent maintains a population of programs that are tested against the environment, evaluated and modified according to the evolutionary strategy described in section \ref{sec:es}. This process happens while the agents interact with the environment. We are basically testing existing generation-based evolutionary algorithms in on-line scenarios. The behaviour execution and evolution global process is depicted in algorithm \ref{alg:gencontevo}. Note that the component that values in this algorithm is the behaviour selection procedure which returns the next program to be used according to certain criteria. This component is described in section \ref{sec:behaviour_selec}.
\begin{algorithm}
\caption{The On-line Generation-Based Evolutionary Algorithm with CGP}
\label{alg:gencontevo}
\algsetup{indent=2em}
\begin{algorithmic}
\vspace{0.5em}
\STATE $P$ \COMMENT{current behavior population}
\STATE $steps \leftarrow 0$
\STATE $\tau \leftarrow t$ \COMMENT{t is the step duration of the evaluation cycle }
\STATE $s_p$ \COMMENT{$s_p$ is the current program selector}
\STATE $p$  \COMMENT{$p$ is the current program being executed}
\STATE $e_p$ \COMMENT{current energy value for program $p$}
\STATE $E_p \leftarrow \emptyset$ \COMMENT{$e_p$ stores a list of energy samples ($e_p$) for the current program $p$}
\STATE $s_p.load(P)$  \COMMENT{loads the current program population into the selector}
\\ \hrulefill 

\IF { ($steps$ $mod$ $\tau = 0$)} 
\STATE $p \leftarrow s_p\text{.next\_program}()$
\ENDIF
\STATE run\_program($p$)
\STATE $\Delta e_p \leftarrow $ energy\_update($p$)
\STATE $e_p \leftarrow e_p + \Delta e_p $
\STATE $E_p \leftarrow \{ E_P, e_p\}$
\IF{($(steps > 0) \text{ AND } (steps \text{ mod } \tau = 0)$)}
	\STATE $s_p$.eval( $\overline{E_p}$ )
	\IF{($s_p\text{is\_done()}$)}
	\STATE $P' \leftarrow \text{breed next generation} (P)$
	\STATE $P \leftarrow P'$
	\STATE $s_p\text{.load}(P)$	
	\ENDIF
\ENDIF

\STATE $steps \leftarrow (steps + 1)$

\end{algorithmic}
\end{algorithm}

\subsection{Behaviour Selection Mechanisms}
\label{sec:behaviour_selec}
We test the generation-based evolutionary strategies algorithm using two different behaviour selection settings. In the first case, similarly to \cite{nordin1997line,Whiteson2006}, we continuously evolve the program population. Each agent start by testing each individual in its environment for a number of steps, after this first round the next programs to be tested are selected according to two different methods:

\begin{itemize}
\item \textbf{Probabilistic Sampling: } each program is selected in a round robin manner so that each program is picked the same number of times to be tested under different conditions, created from the previous program test over the environment \cite{nordin1997line}.
\item \textbf{Interval Estimation Selection: }we compute a $(100-\alpha)\%$ confidence interval for the value of each available program. The agent always takes the program with the highest upper bound. This favours the exploration of programs with high estimated performance or the ones which are most promising but uncertain \cite{kaelbling1993learning,Whiteson2006}.
\end{itemize}

Finally, and in a progressive deepening manner~\cite{antunes2007}, we use both previous mechanisms, but \emph{only if needed}: a program is kept running if its doing well, and it changes, not always for the best, according to a random uniform probability or interval estimation.

In this second method, each agent maintains a virtual energy level that reflects the rewards it receives during the execution of its current behaviour similar to the work in \cite{Elfwing2005}. Each behaviour is evaluated starting with the same energy level and this energy is increased or decreased according to how well the behaviour is doing in terms of its current driving force (current fitness function). In summary, at each step in the simulation, each agent updates its energy value according to:

\begin{equation}
\Delta E(p) = R(p) \cdot \mu E
\end{equation}

\noindent Where $p$ is the current program being run, $R(p) \in [-1,1]$ is the current reward given to the program execution and $\mu E \in [0,1]$ is the energy update constant which dictates how fast an agent depletes or increases its current energy level. This serves two purposes: the evaluation of how well the agent is doing; and the regulation of program evolution, by switching to another program based on how much energy was lost. The fitness function $f(p)$ used in the underlying evolutionary algorithm to evaluate each program $p$ is computed by averaging energy levels, sampled at regular intervals during $\tau$ \textit{simulation steps}.  

The evolutionary strategy uses the fitness evaluations to select the parent individuals used to breed a new generation as described in section \ref{sec:es}. The interval estimation selection mechanism also uses this fitness to choose which program are is to be tested in the environment at the end of each evaluation cycle $\tau$. 

\subsection{Simulation Metrics}
In this paper we present results related to \textit{phenotype fitness}, \textit{genotype diversity}, and \textit{cumulative rewards}. The fitness of the phenotype of the agent is simply the average fitness value given by our underlying function, in this case the Griewank function. This is intended to observe if the agent is behaving in such a way that is gaining as much reward as possible.

Some times is difficult to analyse the overall behaviour of the agent in terms of average fitness over time. To help with this, we also measure the cumulative reward. This is given not by the energy of the agent, but by the sum of all the rewards the agent is given based on the reward computation described in figure \ref{fig:reward}.

Finally, we measure the average genotype diversity for the agent programs being run. To do this we measure the \textit{distance-to-average-point} in genome space. The measure we use was used to alternate between exploring and exploiting behaviour in an evolutionary algorithm proposed in \cite{Ursem2002}. It is also robust with respect to i) the population size (number of genomes being compared in our case), ii) the dimensionality of the problem (number of genes in our CGP programs), and iii) the search range of each of the variables which in this case is within $[0,1]$. The \textit{distance-to-average-point} is defined as:

\begin{equation}
diversity(P) = \frac{1}{\lvert L \rvert \cdot \lvert P \rvert} \cdot \sum_{i=1}^{\lvert P \rvert} \sqrt{\sum_{j=1}^{N}(g_{ij} - \overline{g_j})^2 }
\end{equation}

\noindent where $\lvert L \rvert$ is the length of the diagonal in the search space $S \subset \boldsymbol{R}^N$, $P$ is the population, $N$ is the dimensionality of the problem, $g_{ij}$ is the $j'th$ gene value of the $i'th$ individual of the population, and $\overline{g_j}$ is the $j'th$ value fo the average point $\overline{g}$.

\section{Results and Discussion}
In this section we will present and discuss a series of preliminary exploration scenarios, designed to test various assumptions about the behaviour selection mechanisms. On each experiment performed $30$ independent runs. The number of agents was set to $100$. This is used to test the initial conditions of different agents in the environment since we are not yet explore agent interaction. A mentioned in section \ref{sec:landscape}, the objective is to search for lower ground rather than climbing the peaks as such, the lower the fitness values the better.

\subsection{Continuous Generational Evolution}
In this first experiment we looked at continuous program evolution. In this setting the agents continuously interact with the environment testing the controller programs and evolve using the generation-based evolutionary algorithm describe in section \ref{sec:es}. This implies that in order to evolve the current population of programs, an agent has to test and evaluate each program at least once. A similar setting was used in \cite{nordin1997line}. It was claimed that in spite of the noisy and unfair evaluation (good individuals dealing with hard situations can be rejected in favour of a bad individual dealing with a very easy situation), the good overall programs tend to survive and reproduce in the long term. 

To deal with some of the continuous on-line sparse evaluation problems, we tested the interval estimation method described in section \ref{sec:behaviour_selec}. A similar setting was proposed in \cite{Whiteson2006} to evolve neural networks.  

We observed the effect of two different program selection mechanisms in the agents overall performance, this is, the capacity of identifying and spending more time in good regions. This is useful to establish a baseline of comparison for future work and identifying possible problems, namely, we want to see how generation-based evolutionary algorithms respond to continuous evolution.

\begin{figure}
     \centering
     \subfloat[]{
     \includegraphics[width=0.33\linewidth]{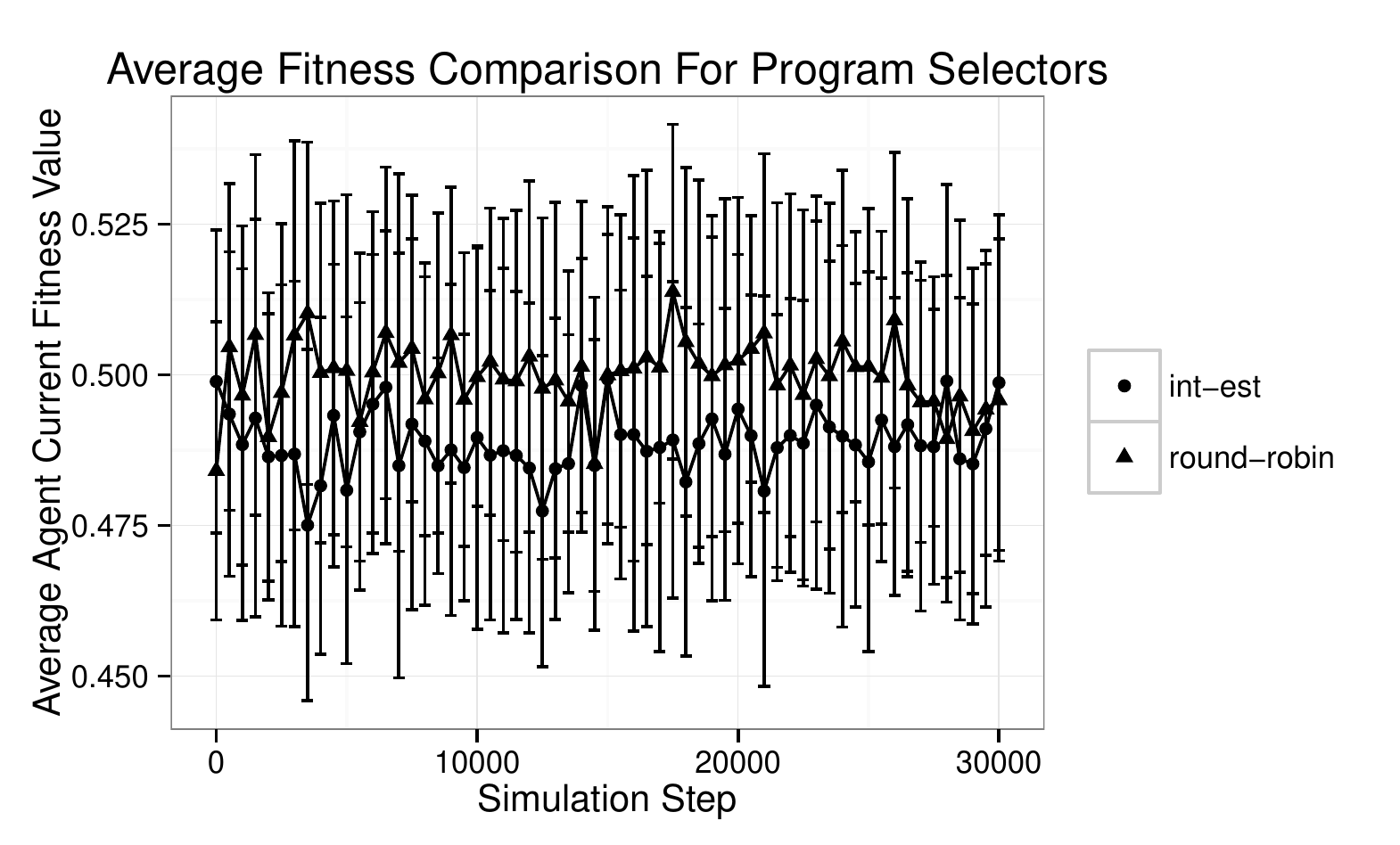}
     \label{fig:exp2_05_f}
     }
     \subfloat[]{
  \includegraphics[width=0.33\linewidth]{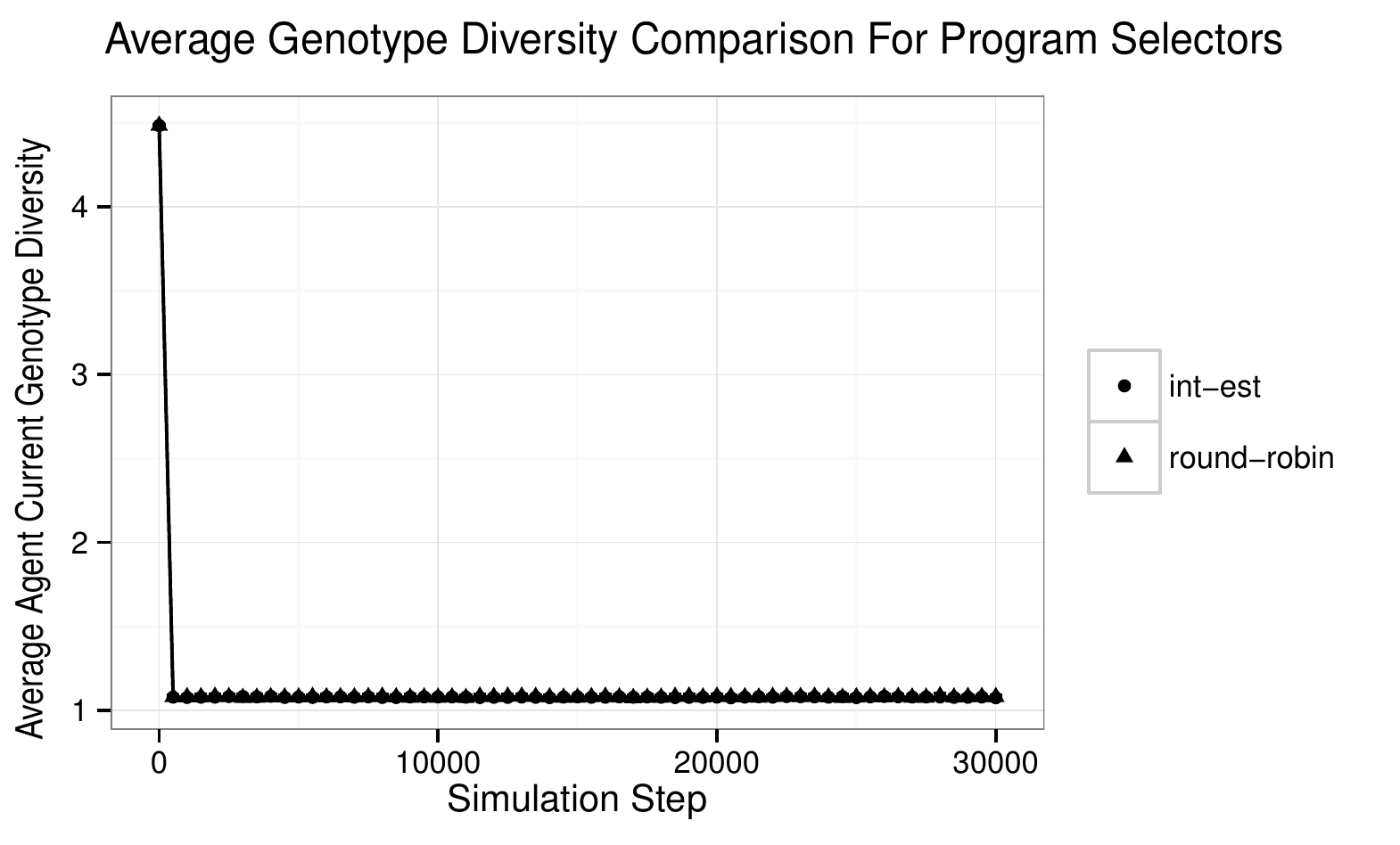}
   \label{fig:exp2_05_d}
     }
      \subfloat[]{
      \includegraphics[width=0.33\linewidth]{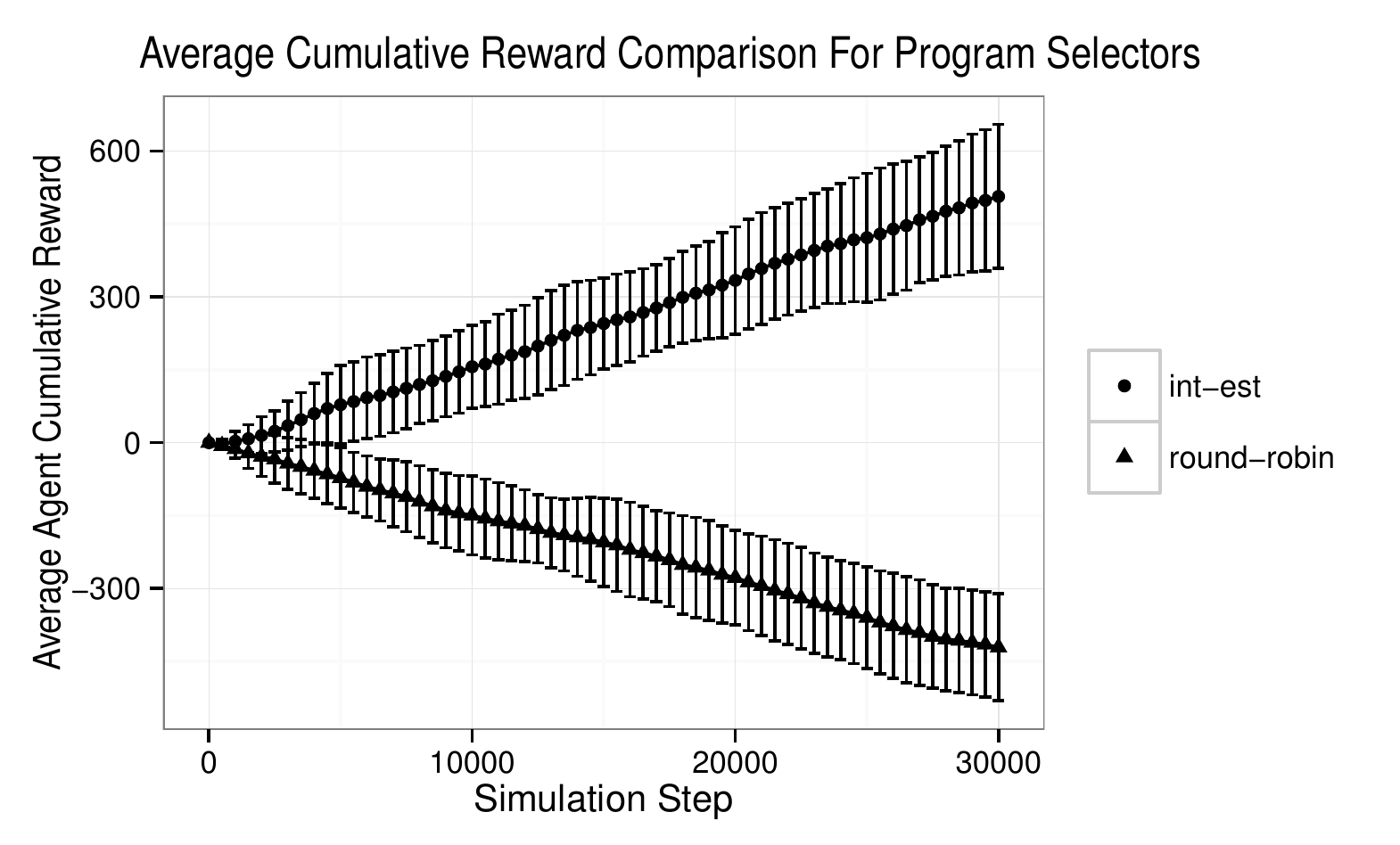}
        \label{fig:exp2_05_c}
          }
     \caption{Comparison between round-robin and interval estimation program selection in terms of average fitness (a), diversity (b), and cumulative reward with the reward threshold set to $0.5$.}
     \label{fig:exp2_05}
\end{figure}

Figure \ref{fig:exp2_05_f} shows that the agents navigated through regions that were close to the fitness value $0.5$. Since the threshold reward was set to $0.5$ the only way to gain energy was to find regions with higher fitness values. In this case, only the interval estimation approach was able to achieve a positive cumulative reward over time (see figure \ref{fig:exp2_05_c}). We can also see that continuous evolution completely ruined genotype diversity, compared to the initial levels.

\noindent Although our approach using CGP and the one in \cite{nordin1997line} are similar, we didn't observe a long term adaptation in continuous evolution. We can only conclude that the claims made in \cite{nordin1997line} are based on the fact that this is possibly highly dependent on the environment,the task being performed, the evolutionary algorithm, and possibly on the representation.

\subsection{Varying Reward Threshold}
In this next experiment we started with lower values of the reward threshold. In this case, we set the reward threshold $R_T$ to $0.5$. As we are trying to lead the agents to lower areas of the landscape, fitness values from $0.5$ to $0$ give a reward value of $0$ to $1$ respectively (see figure \ref{fig:reward}). We compared the two energy-based stabilisation program selector mechanisms (uniform probability and interval estimation re-sampling) with a society of agents that use the energy-based stability rule but does not evolve its program population.

\begin{figure}
     \centering
     \subfloat[]{
  \includegraphics[width=0.32\linewidth]{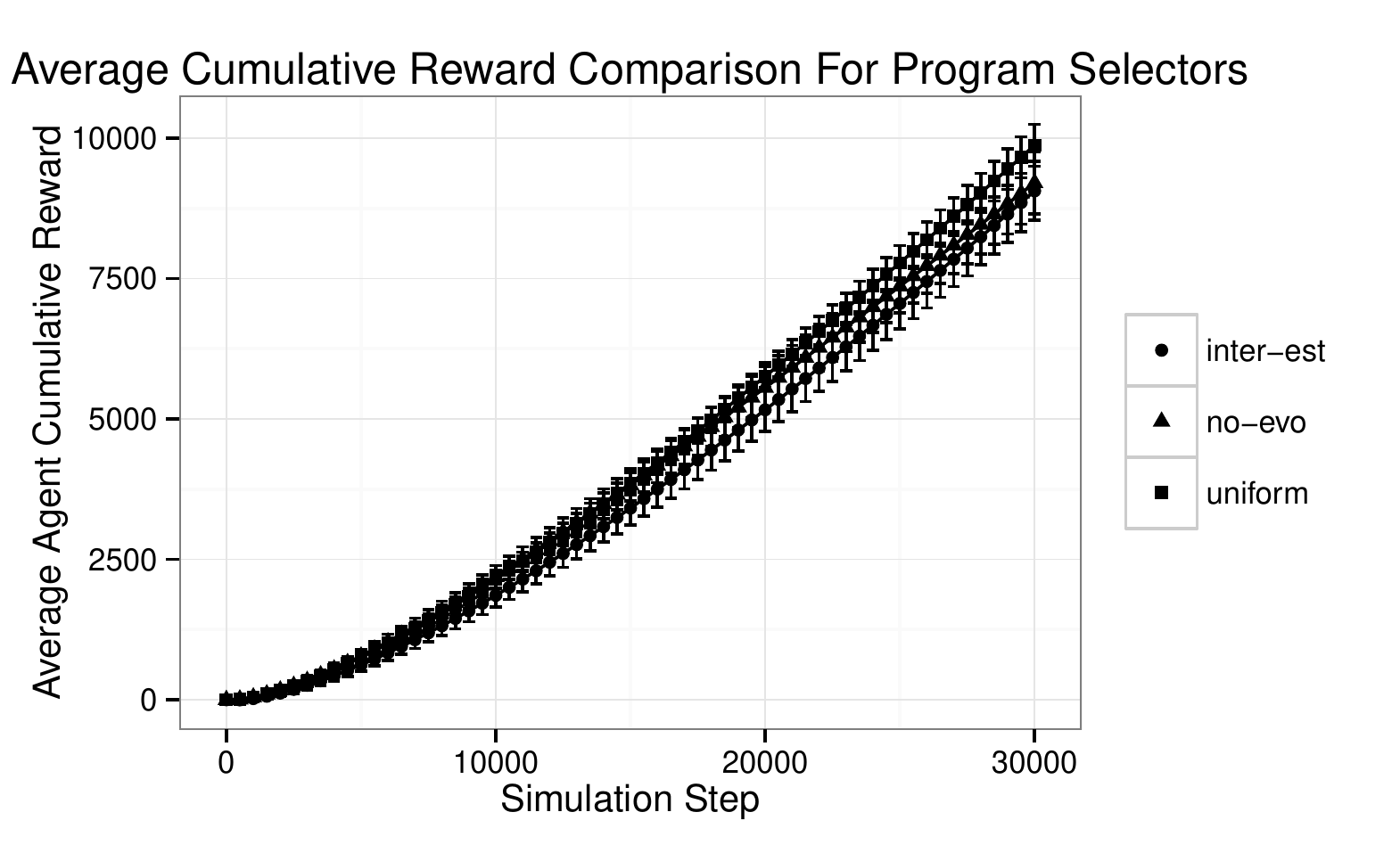}
   \label{fig:exp1_05_c}
     }
     \subfloat[]{
       \includegraphics[width=0.32\linewidth]{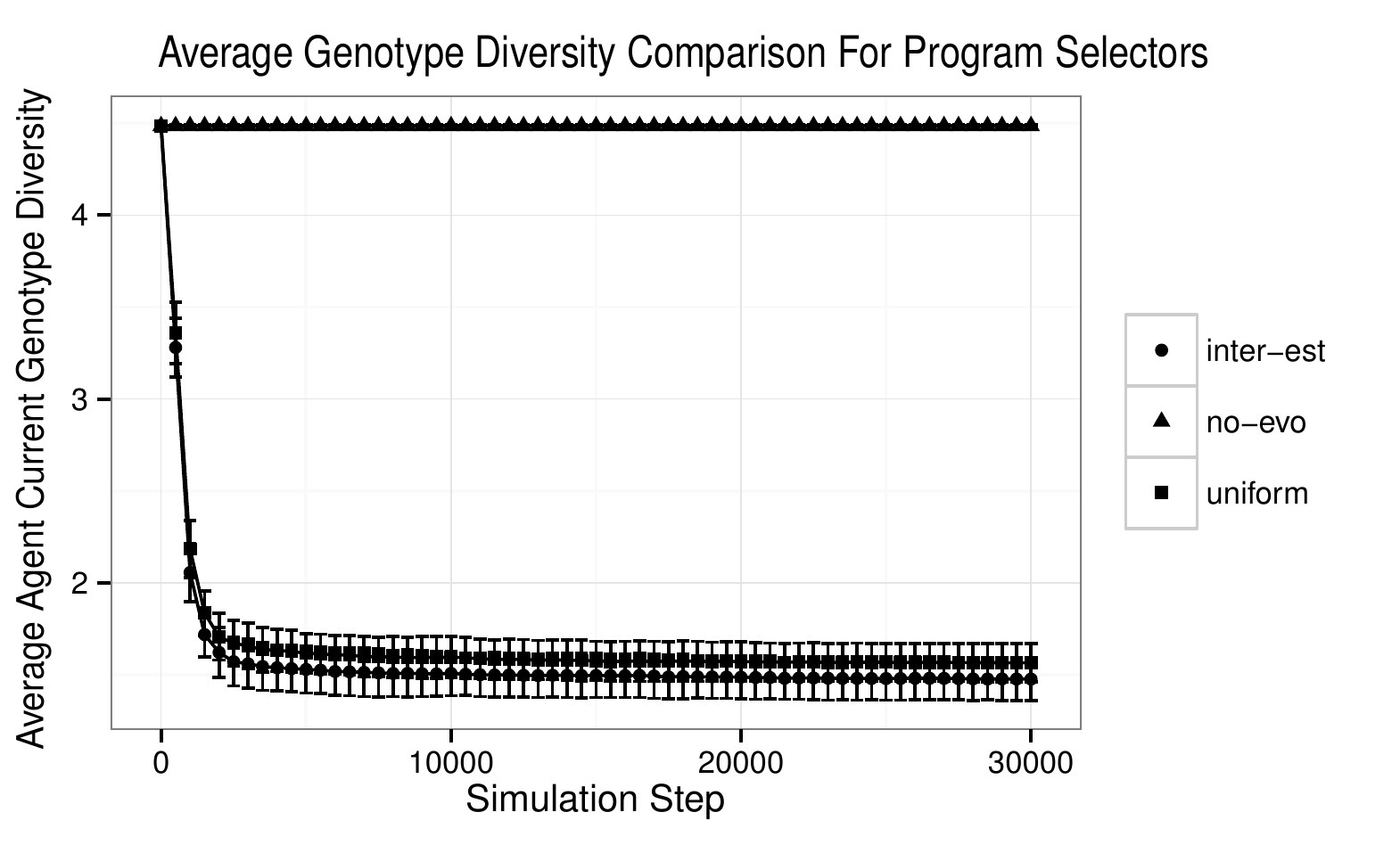}
        \label{fig:exp1_05_d}
          }
            \subfloat[]{
               \includegraphics[width=0.32\linewidth]{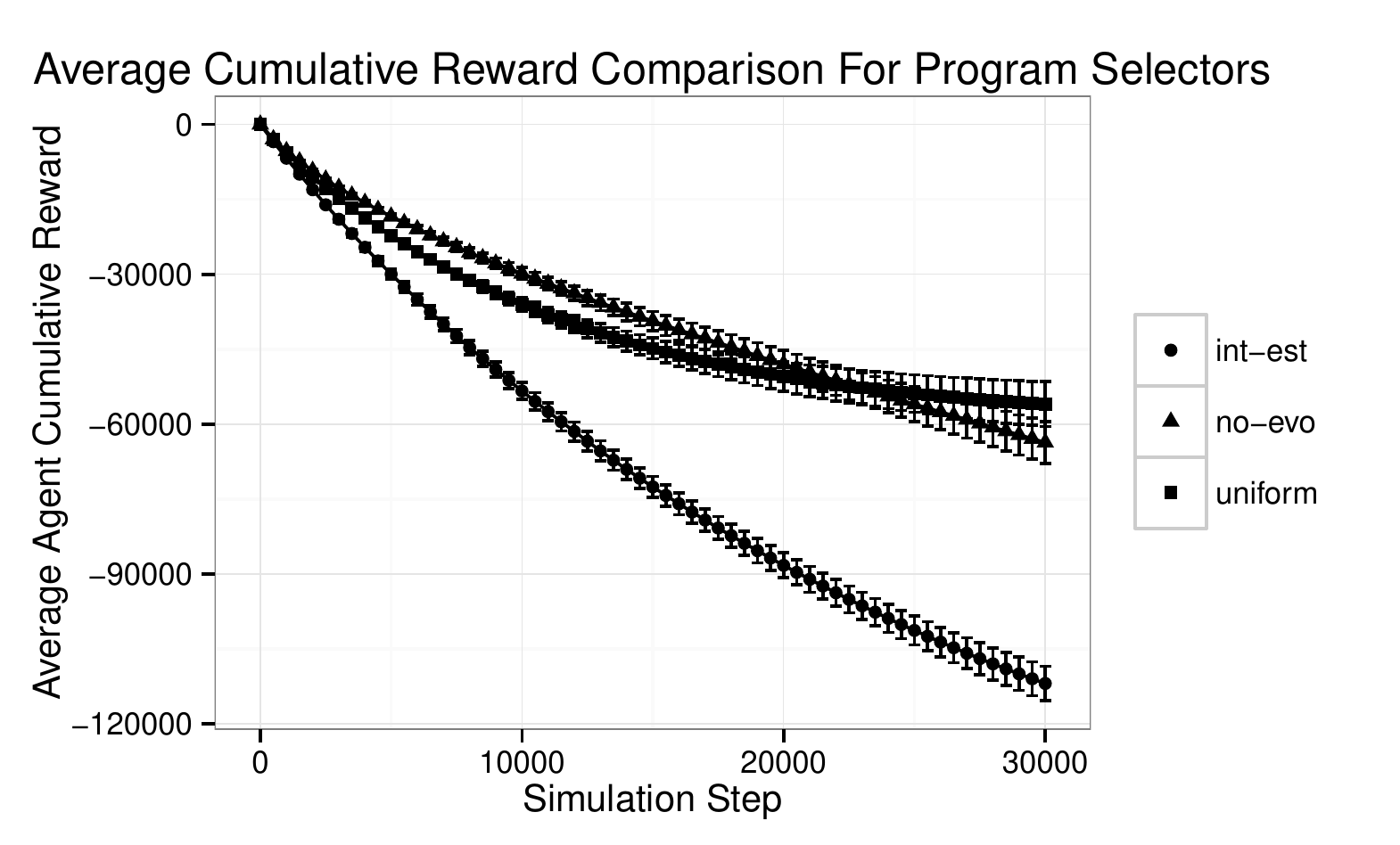}
               \label{fig:exp1_01_c}
               }
     \caption{Average cumulative reward (a) and average diversity (b), for reward threshold $R_T = 0.5$ and average cumulative reward for $R_T = 0.1$ for tree different program selection mechanisms.}
     \label{fig:exp1_05}
\end{figure}

Figure \ref{fig:exp1_05_c} shows the average cumulative reward for an experiment where the threshold reward was set to $0.5$. We can observe that the cumulative reward values are very similar, with the uniform probability selection having a slightly superior cumulative reward.

Our hypothesis here is that the agents that don't evolve their behaviours have set of programs that are diverse enough to contain both behaviour policies that allow them to navigate in the environment and the ones that allow them to remain in one area. If they posses the capabilities to stay in one area, the energy-based stabilization will do the rest and ensure that if they selected an appropriate program, this stays in execution. The approach without evolution was also able to do particularly well in this scenario because the reward threshold is tolerant enough to create an environment with plenty of reward opportunities.

We can see that both the uniform probability and the interval estimation methods (figure \ref{fig:exp1_05_d}) bring down the genotype diversity at the beginning of the simulation and maintain that value throughout the rest of the run while the agents continue to improve. The diversity was slightly higher for uniform probability selection which was confirmed using a t-test with 95\% confidence.

Figure \ref{fig:exp1_01_c} shows a different scenario where the reward threshold was set to $0.1$, we can see that in this case, finding reward regions becomes more difficult and only the uniform probability with energy-based regulation seems to te stabilising its negative cumulative reward.

\subsection{Diversity and Evolutionary Parameters}
Finally, this experiment we are interested in average genotype diversity and fitness response to the parameters of our underlying generational evolutionary algorithm. Here we present the uniform probability selector with energy-based stabilisation and vary the maximum number $n_p$ of nodes one program can have in $n_p \in \{100,200,300,400\}$ and the global mutation rate $m_p \in \{.01, .02, .03, .04\}$. Since we observed an early convergence of diversity (before $5000$ steps) in previous experiments (figure \ref{fig:exp1_05_d}), we set the maximum number of steps to 5000.

\begin{figure}
     \centering
     \subfloat[]{
     \includegraphics[width=0.40\linewidth]{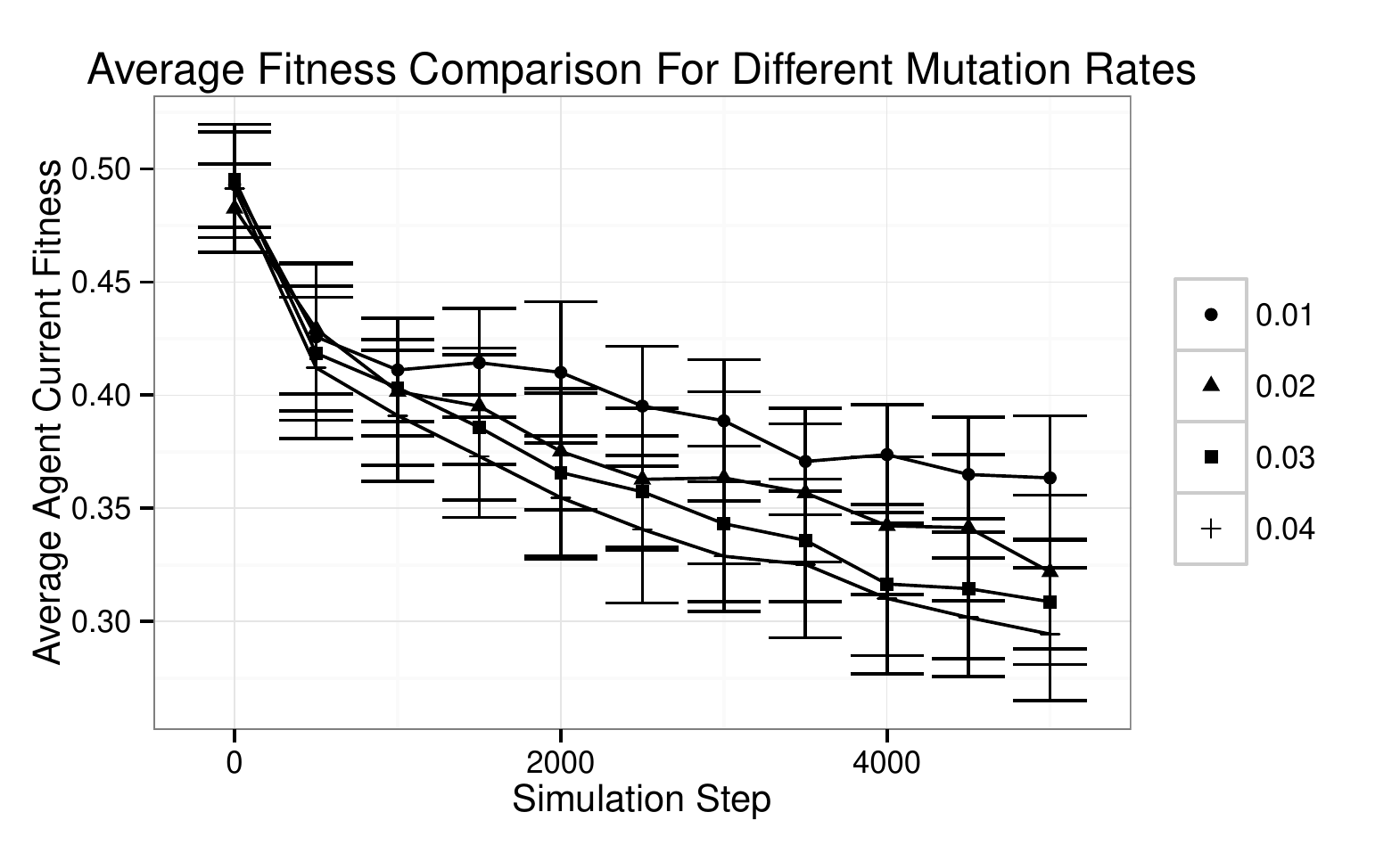}
     \label{fig:exp4_f_m}
     }
     \subfloat[]{
  \includegraphics[width=0.40\linewidth]{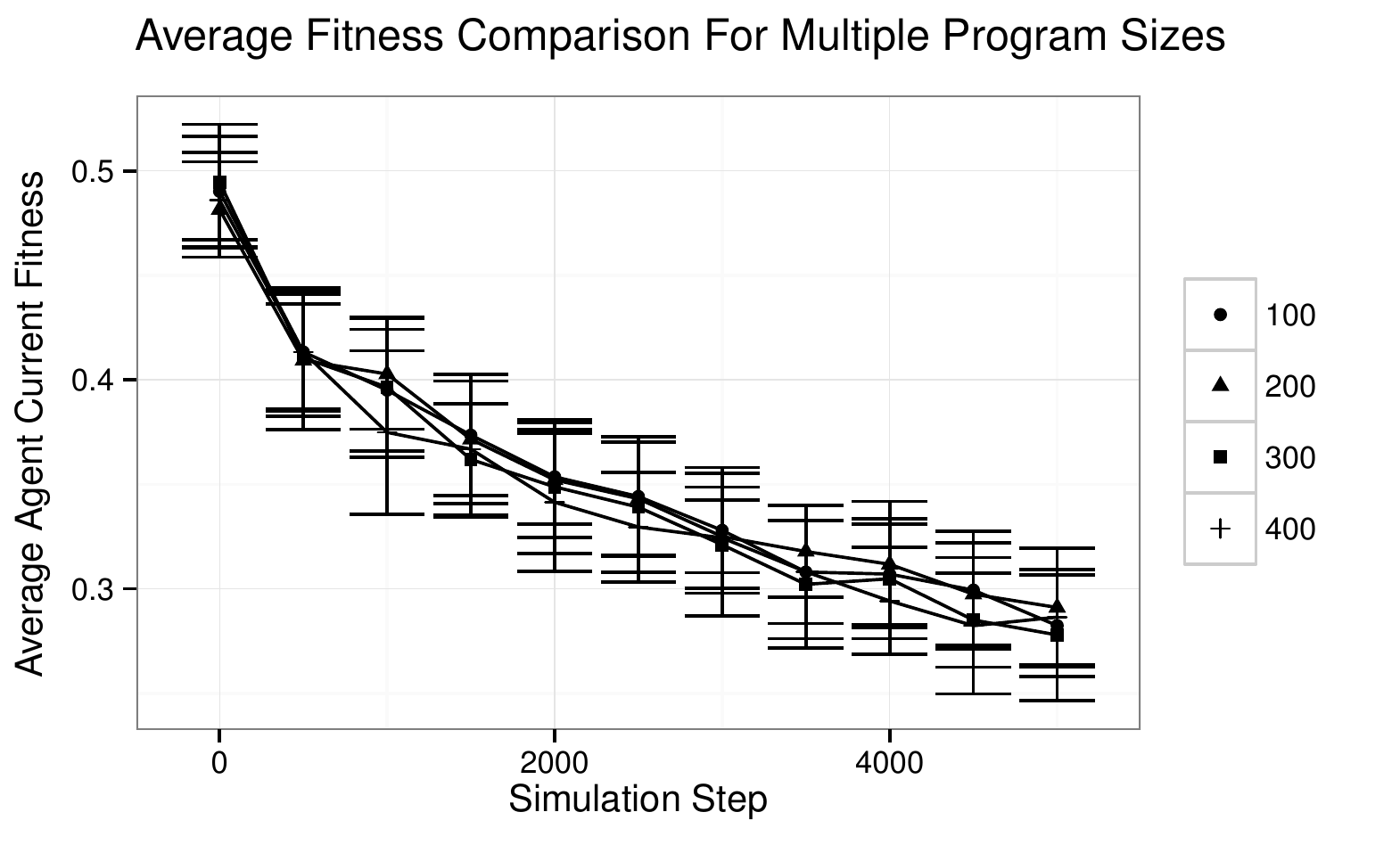}
   \label{fig:exp4_f_n}
     }\\
      \subfloat[]{
       \includegraphics[width=0.40\linewidth]{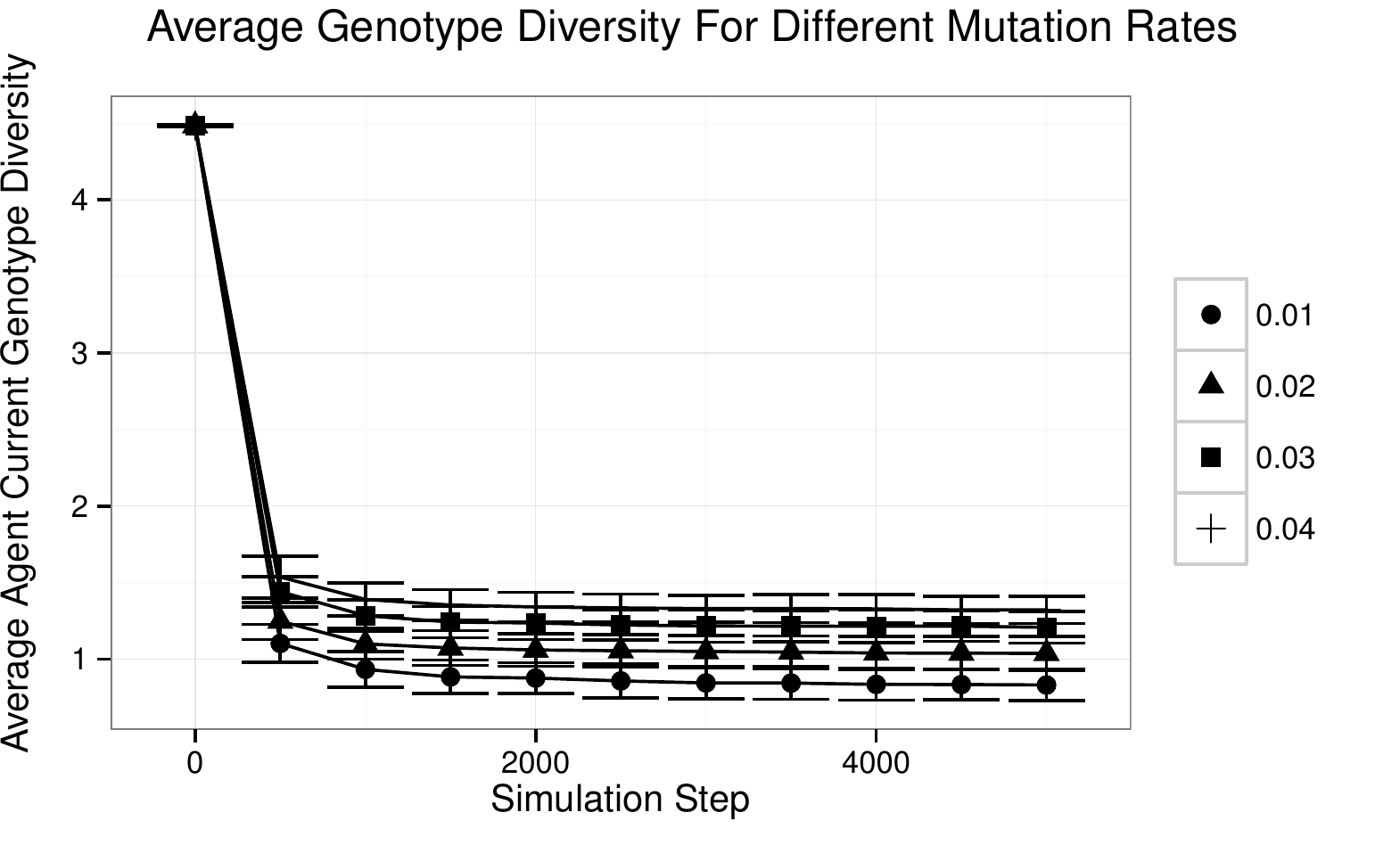}
        \label{fig:exp4_d_m}
          }
            \subfloat[]{
            \includegraphics[width=0.40\linewidth]{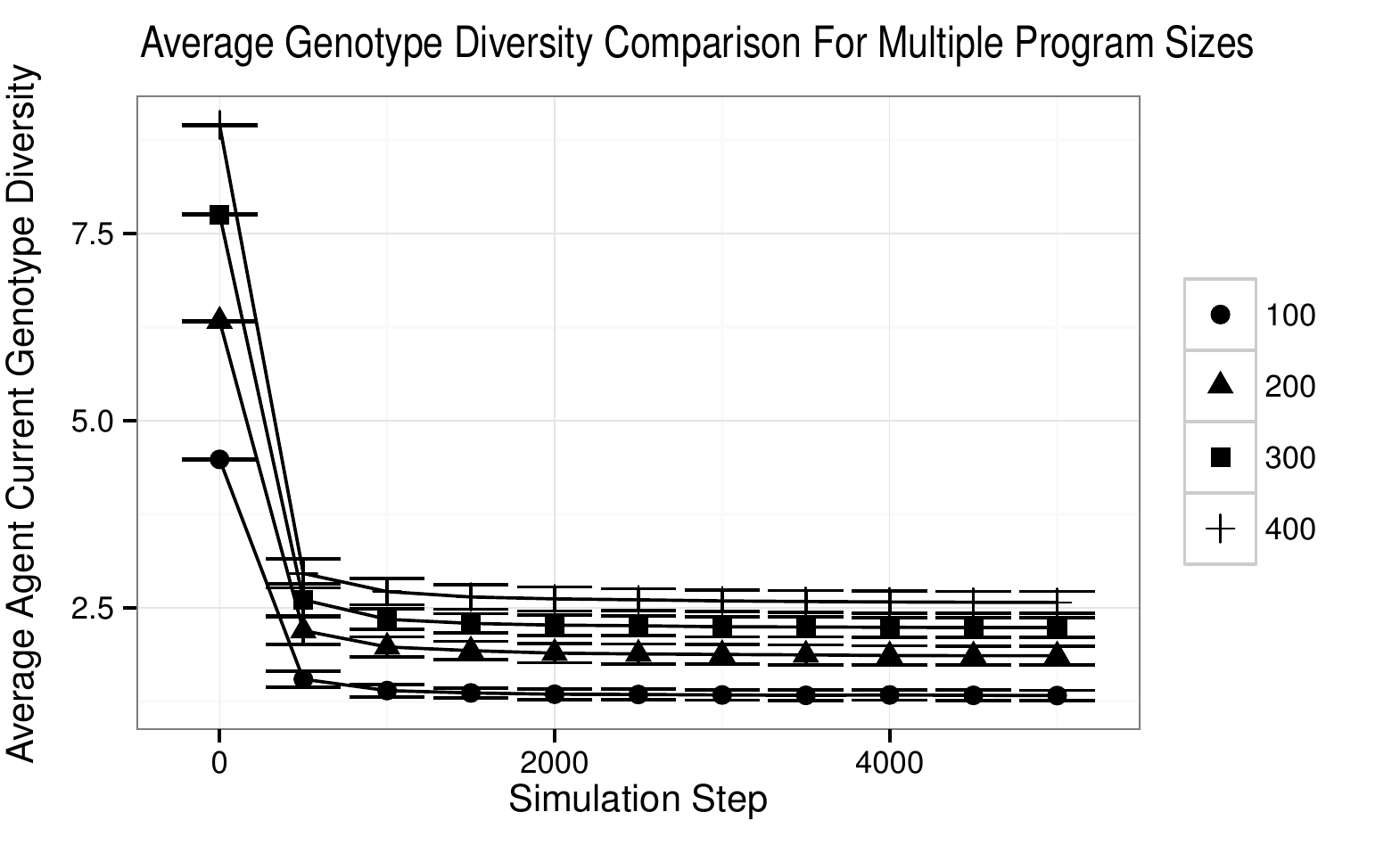}
             \label{fig:exp4_d_n}
               }
     \caption{Average fitness given by the \textit{Griewank} function (a,b) and average genotype diversity (c,d) with reward threshold set to $0.01$. In (a,c) the number of nodes is fixed in $100$, in (b,d) the mutation rate is fixed in $0.04$}
     \label{fig:exp4_01}
\end{figure}

For this selector mechanism, higher mutation rates lead to better average fitness over time (figure \ref{fig:exp4_f_m}) as well as higher diversity values (figure \ref{fig:exp4_d_m}). With the mutation rates fixed in $0.04$, we can see that, while the number of maximum nodes does not affect the average fitness over time, bigger CGP programs display higher levels of genotype diversity. This confirms that although the CGP program representation is very sensitive to mutations in terms of phenotype expression, higher mutation rates seem to be beneficial to maintain diversity, mutations are possibly attenuated by inactive program nodes. In future, we will also analyse the phenotype diversity and its relationship with the active and inactive nodes in the program in a similar on-line domain.

\section{Conclusion and Future work}
In this paper, have tested a generation-based evolutionary strategy to evolve cartesian programs in an on-line scenario. This is a first step into a series of studies involving evolutionary computation to create truly autonomous agents with self motivations. We chose the evolutionary programming path as it allows for behaviours to recursively and reflectively build
on themselves, given the adequate operators. So, the potential for incremental complexity is unlimited. In future work, we will test different scenarios where agents will evolve not only in response to adaptive exploration of fitness functions that the environment offers, but also in response to social drives that adapt controller programs to imitate or compete against a target mediators.

Our purpose is to address both issues in evolutionary learning methods for interacting multi-agent systems, but also to provide a more realistic account of in individual choice, not focused on optimising given measures. We claim that this can be beneficial in areas such as social simulation where more trustworthy scenarios can be
run, possibly to rehearse and help the deployment of policies.

\bibliography{main}
\bibliographystyle{splncs03}

\end{document}